\pdfoutput=1

\documentclass[letterpaper, 10pt, conference]{ieeeconf}      %

\IEEEoverridecommandlockouts                              %

\let\labelindent\relax                                    %

\usepackage{graphicx}            %
\usepackage{color}
\usepackage{amsmath}             %
\usepackage{amssymb}             %
\usepackage{amsfonts}            %
\usepackage{enumitem}            %
\usepackage{multirow}            %
\usepackage{siunitx}             %
\usepackage{url}                 %
\usepackage{xspace}              %
\usepackage[T1]{fontenc}         %
\usepackage[hidelinks]{hyperref} %
\usepackage{balance}
\usepackage[bottom]{footmisc}
\usepackage{xfrac}
\usepackage{tensor}
\usepackage{todonotes}

\usepackage[firstpage=true]{background}
\newcommand\copyrighttext{%
\parbox{\textwidth}{
\footnotesize
In Proceedings of IEEE International Conference on Robotics and Automation (ICRA), Paris, France, May 2020
}}

\SetBgContents{\copyrighttext}
\SetBgScale{1}
\SetBgColor{black}
\SetBgAngle{0}
\SetBgOpacity{1}
\SetBgPosition{current page.north}
\SetBgVshift{-0.8cm}

\usepackage[bottom]{footmisc}

\graphicspath{{images/}}
\DeclareGraphicsExtensions{.pdf,.png,.jpg,.jpeg}

\DeclareMathOperator{\acos}{acos}

\providecommand{\norm}[1]{\lVert#1\rVert}

\newcommand{\mypm}{\mathbin{\mathpalette\@mypm\relax}}

\makeatletter
\newcommand{\amsray}{%
\mathpalette {\overarrow@\rayfill@}}
\def\rayfill@{\arrowfill@{\mkern4mu\mapstochar\relbar}\relbar{\mkern 4.08mu}}%
\makeatother

\newcommand{\seclabel}[1]{\label{sec:#1}}

\newcommand{\figlabel}[1]{\label{fig:#1}}
\newcommand{\tablabel}[1]{\label{tab:#1}}
\newcommand{\eqnlabel}[1]{\label{eqn:#1}}

\newcommand{\secref}[1]{Section~\ref{sec:#1}\xspace}

\newcommand{\figref}[1]{Fig.~\ref{fig:#1}\xspace}
\newcommand{\tabref}[1]{Table~\ref{tab:#1}\xspace}
\newcommand{\eqnref}[1]{(\ref{eqn:#1})\xspace}

\newcommand{\iguhop}{igus\textsuperscript{\tiny\circledR}$\!$ Humanoid Open Platform\xspace}

\newcommand{\cpp}{C\texttt{\nolinebreak\hspace{-.05em}+\nolinebreak\hspace{-.05em}+}\xspace}

\title{\LARGE \textbf{Fast Whole-Body Motion Control of Humanoid Robots\\with Inertia Constraints}}

\author{Grzegorz Ficht and Sven Behnke%
\thanks{All authors are with the Autonomous Intelligent Systems (AIS) Group, Computer Science Institute VI,
        University of Bonn, Germany. Email: {\tt\small ficht@ais.uni-bonn.de}. This work was partially
        funded by grant BE 2556/13 of the German Research Foundation (DFG).}}

\begin{document}

\bstctlcite{IEEEexample:BSTcontrol}

\maketitle
\thispagestyle{empty}
\pagestyle{empty}

\begin{abstract}

We introduce a new, analytical method for generating whole-body motions for humanoid robots, which approximate 
the desired Composite Rigid Body~(CRB) inertia. Our approach uses a reduced five mass model, where four of the masses are 
attributed to the limbs and one is used for the trunk. This compact formulation allows for finding an analytical 
solution that combines the kinematics with mass distribution and inertial properties of a humanoid robot. 
The positioning of the masses in Cartesian space is then directly used to obtain joint angles with relations based on
simple geometry. Motions are achieved through the time evolution of poses generated through the desired foot positioning 
and CRB inertia properties. As a result, we achieve short computation times in the order of tens of microseconds. This makes
the method suited for applications with limited computation resources, or leaving them to be spent on higher-layer tasks 
such as model predictive control. The approach is evaluated by performing a dynamic kicking motion with an \iguhop robot.

\end{abstract}
\section{Introduction}

Control of humanoid robots is a challenging task, as they belong to a class of high-dimensional, underactuated systems. 
Full mathematical formulations containing all of the precise physical details about the system are highly complex.
As every moving body part contributes to the overall dynamic behaviour, motion planners are required to traverse 
a broad search space, resulting in long solve times. As a solution, several simplified models have been used, 
which approximate the general dynamics of the system. 

Single mass pendulum models are quite popular in this regard, 
as they allow to characterise the linear momentum of the system through a trajectory of the Center of Mass~(CoM). 
A linear simplification of the inverted pendulum~\cite{kajita20013d} has been widely used in bipedal walking~\cite{kajita2003biped}.
Through such simple models, even comprehensive and efficient control strategies can be achieved~\cite{missura2013omnidirectional}.
The beneficial simplicity is accompanied with limitations of not including the angular momentum and a constant CoM height.
Three mass models~\cite{takenaka2009real}\cite{shimmyo2012biped} were used to model leg swing dynamics and accordingly alter the 
Zero Moment Point~(ZMP)~\cite{vukobratovic2004zero} for stable bipedal walking. Once the dynamic trajectories have been computed, 
the joint states are then usually computed using numerical inverse kinematics to satisfy the desired CoM.
A generalisation for the centroidal dynamics~\cite{orin2013centroidal} was proposed by Lee et al.~\cite{lee2007reaction}~ with the Reaction Mass Pendulum~(RMP),
where six equal masses sliding on three actuated rails shape the total inertia of the robot. The masses are abstract and do not 
relate to the joint angles in a direct way. As a result, the mapping between inertia and whole-body pose is not unique.

More general whole-body controllers that address a number of criteria have also been developed. One of these---the Resolved Momentum Control
framework---unified control over linear and angular momenta~\cite{kajita2003resolved} and generated naturally looking kicking and walking motions.
To solve broader and less-structured problems than biped locomotion, task-based optimisation methods are commonly employed. Through their usage, various
multi-task~\cite{sentis2005synthesis}~\cite{del2014prioritized} problems can be solved. Depending on the constraints and details
included in the model, the solver can take up to hours of computation~\cite{lengagne2013generation}. Recent works have been targeted at reducing solve 
times through simplifying the dynamics~\cite{dai2014whole} or through their more efficient computation~\cite{wensing2016improved}.

In our previous work~\cite{ficht2018online}, we have shown that whole-body poses that satisfy a given CoM placement can be computed 
using a five mass model and relatively simple geometry with a triangle mass distribution. At the same time, Arreguit et al. 
\cite{arreguit2018fast} used a five mass model with stretchable limbs for simplified dynamics in motion planning tasks to 
achieve solve times in the order of milliseconds. The five mass model efficiently encodes the combination of limb swing 
dynamics and their kinematics (\figref{inertia_teaser}). As we show in this work, it also provides insight in determining pose reachability constraints and 
allows for generating whole-body motions analytically in microseconds. The contribution of this paper is a novel, computationally 
efficient method for controlling humanoid robots through setting a combination of foot-relative CoM position and rotational inertia.
A video is available\footnote{\url{http://ais.uni-bonn.de/videos/ICRA_2020_Ficht}}.

\begin{figure}[!t]
\centering{\includegraphics[width=0.99\linewidth]{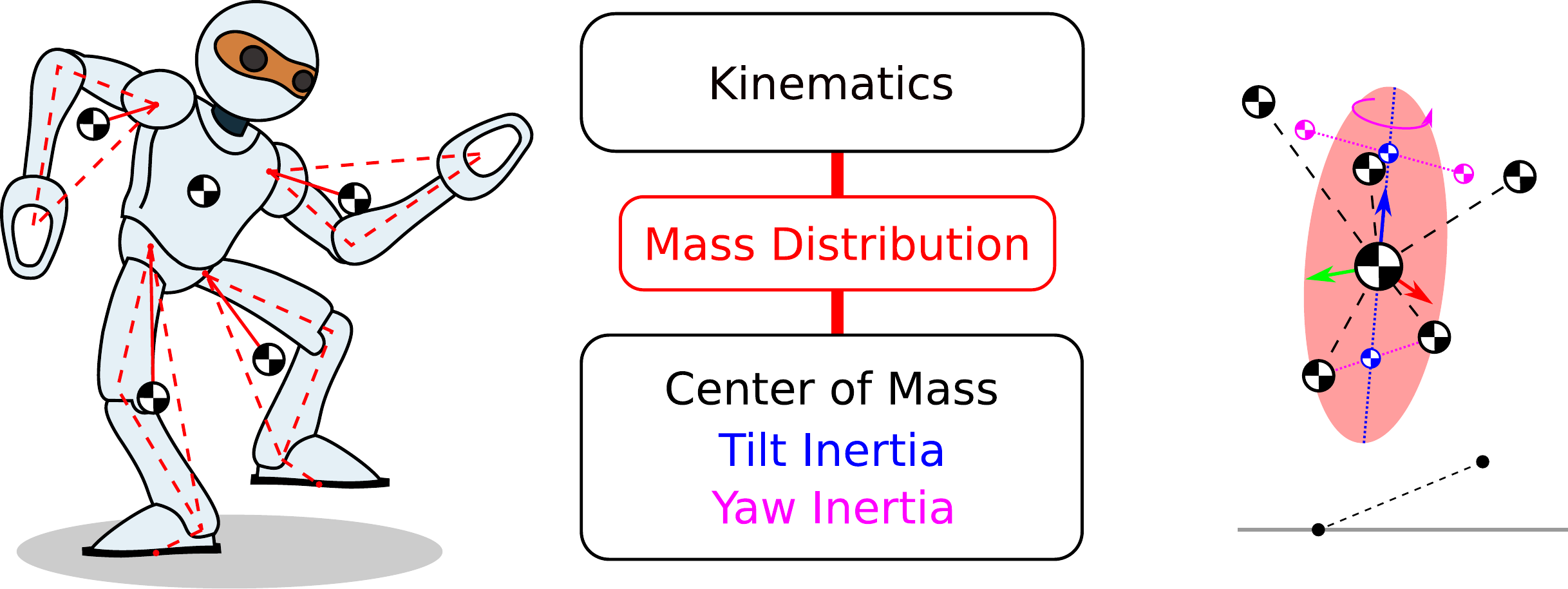}}
\caption{Approximation of a humanoid robot with a five mass model, efficiently encoding the kinematics and inertial properties of the system.
This formulation allows for fast, analytic generation of whole-body motions.}
\figlabel{inertia_teaser} 
\vspace{-2ex}
\end{figure}
\section{Reduced Five Mass Model Description} \seclabel{model_reduction}

We describe a humanoid robot as a tree-structured rigid body consisting of five point-masses with a non-uniform mass distribution.
A 6 degree of freedom(DoF) floating base frame $\Sigma_B$ is attached to the pelvis of the torso, between the hip joint origins $\mathbf{h_m}$.
We consider the head a part of the torso, and place this combined mass $m_t$ at a configurable offset from the base. This leaves the head unconstrained, %
which is desirable in vision systems. The remaining four links are attached to the torso and represent the (left and right) leg $m_{ll},m_{rl}$ and arm $m_{la},m_{ra}$,
which originate at the hip and shoulder joints respectively. The centroidal properties of the system, i.e. CoM position $\mathbf{m_c}$ and orientation
of the principal axes of inertia $\mathbf{R_I}$ are combined in frame $\Sigma_C$. All movable joints such as hips, knees, ankles, shoulders, elbows and wrists
possess their own frame. Furthermore, two foot frames $\Sigma_{FL},\Sigma_{FR}$ are placed at $\mathbf{f_l},\mathbf{f_r}$ and represent the centers of the foot polygon. For the convenience of further calculations, all position vectors are expressed in the CoM frame, i.e. $\mathbf{m_c} = (0,0,0)$, while orientations follow a right-handed coordinate system.%

\begin{figure}[!t]
\centering{\includegraphics[width=0.99\linewidth]{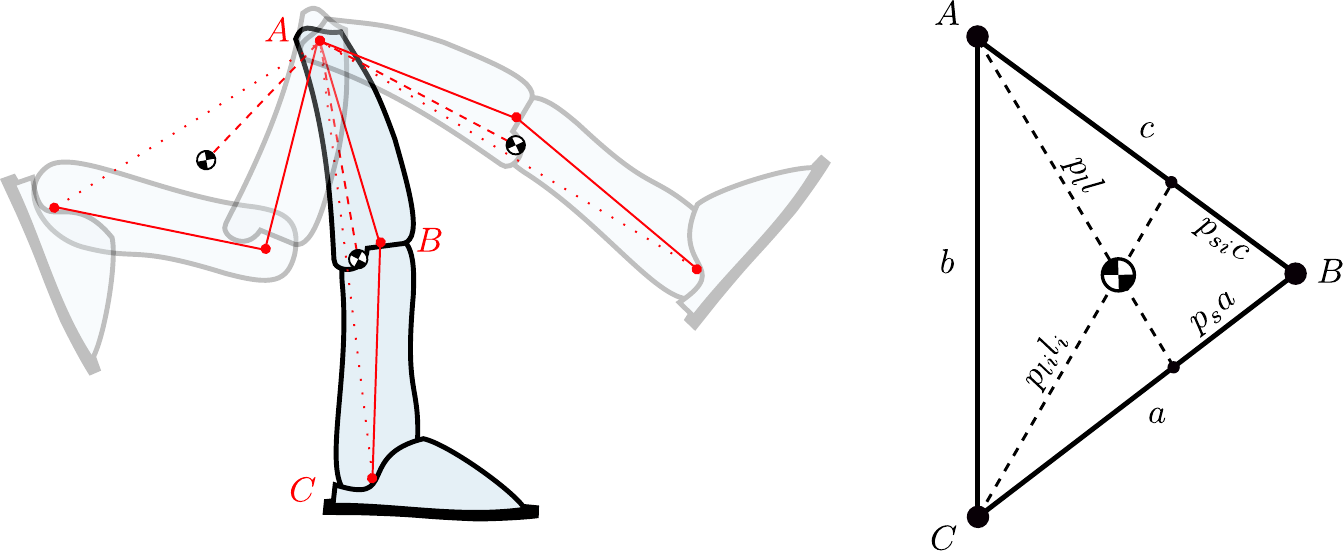}} %
\caption{Limb kinematics in relation to mass positioning. Using a triangle mass distribution---parameterised in $P$--- a direct relation between limb mass movement and joint angles is achieved.}
\figlabel{triapprox}
\vspace{-2ex}
\end{figure}

The limb mass position is tied-in with the kinematics through a triangle approximation. Two sides of the triangle are of known, constant length~($c$,$a$) and represent
the actual upper and lower leg and arm links of the robot. The third side corresponds with the variable limb extension $b$, dependent on the knee or elbow bending angle.
In a limb, two \textit{mass distribution parameters} are used to determine the relative position of the mass in the triangle with respect to the origin:
\begin{equation}
P = (p_s,p_l)\in[0,1]. 
\eqnlabel{parameters}
\end{equation}%
The \textit{side distribution parameter} $p_s$ is used to section the lower link of the limb. The center of mass is located along the vector formed 
between this intersection point and the limb origin. The \textit{length distribution parameter} $p_l$ represents the ratio between the CoM distance 
to the length of this vector. A triangle with uniform density, for example, has $p_s = \frac{1}{2}$, $p_l = \frac{2}{3}$, with the vector between the origin and 
CoM being a median. By assigning $P$ values for each limb, a one-to-one mapping is formed through which joint values can be obtained from mass positioning, 
and vice versa. Additionally, inverse distribution parameters $P_i = (p_{si},p_{li})$ can be calculated through simple intersections. 
These can then be used to find the configuration of a limb depending on the mass distance from the wrist or ankle. 
The connection between mass and limb movement is depicted on \figref{triapprox}. 
Although the mass of the feet and hands is included in the total limb mass, it is assumed that their orientation does not influence 
the final mass placement in the triangle. This is due to their limited range of motion and low weight relative to the rest of the limb. 
As shown in previous work~\cite{ficht2018online}, such a relationship between the mass distribution and limb kinematics is precise, 
given that $P$ was determined with sufficient accuracy.

This five link formulation is more complex than typically used single point-mass models, however it possesses beneficial properties 
which simpler models are unable to capture. The main advantage is the ability to describe the robot as an equivalent Composite Rigid Body~(CRB) 
with varying rotational inertia. This allows to characterise the centroidal dynamic properties of the system, and control the linear and angular momentum.
In comparison to the already mentioned Reaction Mass Pendulum~\cite{lee2007reaction}, the masses are not abstract and have a direct connection to the physical system. 
Not only do they describe the non-uniform mass distribution, but also control the kino-dynamic properties. Meaning that a change in the positioning of a 
single mass, affects the CRB inertia and the limb configuration in a clear and unique way.

\section{Analytic Whole-Body Control}  \seclabel{wbcontrol}

Using the model described in the previous section, we propose a novel method for whole-body control.
We capitalise on the relationships of mass distribution and kinematics, and derive analytic solutions based on geometry.
For the pose generation process, the desired foot placement and inertial properties are used. 
This input, combined with the physical system properties is then used to compute target joint values.
The complete process flow of generating a whole-body pose is presented in \figref{flowchart}.

\begin{figure}[!t]
\centering{\includegraphics[width=0.99\linewidth]{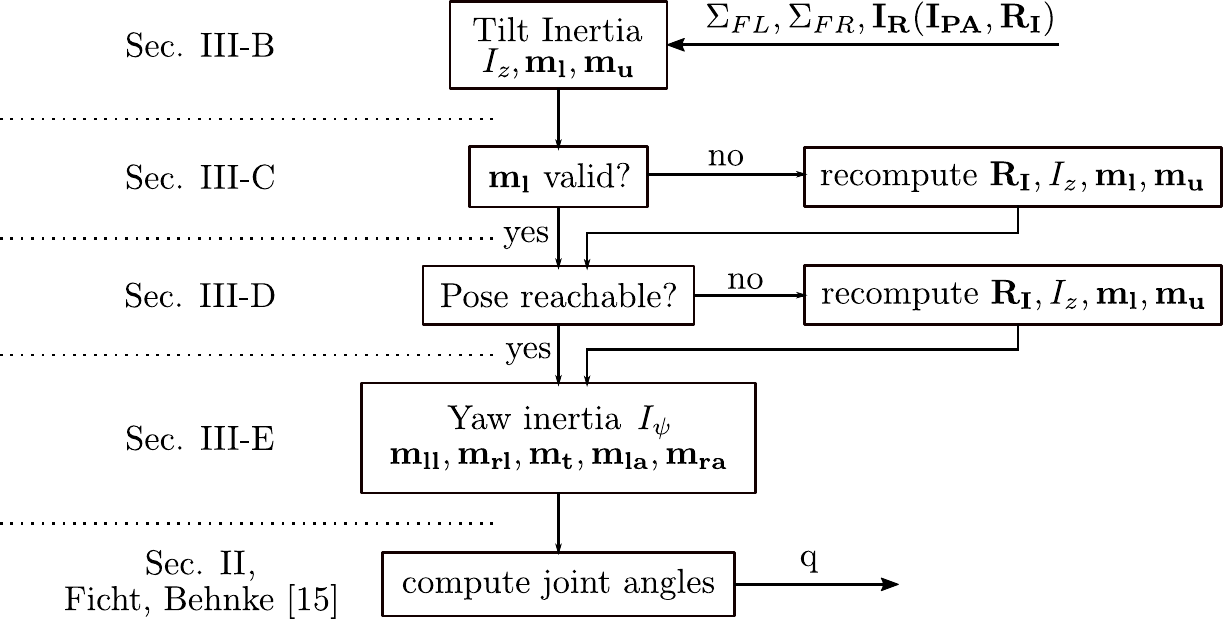}} %
\caption{Complete process flow of generating a whole-body pose using the presented approach.}
\figlabel{flowchart}
\vspace{-2ex}
\end{figure}

\subsection{Constraint inclusion} %

A number of constraints are considered which act as the system input. The highest priority is given to CoM placement, which is constrained
by the foot positions. The inertial properties of the multibody are considered to be of lower priority, where the inertia orientation takes precedence
over the principal moments in some cases. As we show in \secref{footplacement}, foot placement is the main constraint defining factor with respect to both 
the CoM and inertia. Lastly, when a pose has been calculated that respects the possible limits, an optional constraint on the trunk orientation can be employed.
Although a hierarchy of the constraints exists, it is not achieved through weight assignment. As our method does not rely on a global task optimisation function,
the priority of the constraints is a result of the procedural approach to generating the whole-body pose. Kinematic constraints are included through the 
triangle approximation. %

\subsection{Tilting inertia} \seclabel{dumbbell}

The rotational inertia of rigid bodies $\mathbf{I_R}$ taken at the CoM, has the form of:
\begin{equation}\eqnlabel{principalmoments}%
\begin{aligned}
\mathbf{I_R} = \mathbf{R_I}\mathbf{I_{PA}}\mathbf{R_I^T} = \mathbf{R_I}\begin{bmatrix}I_{xx} &0 &0\\0 &I_{yy} &0\\0 &0 &I_{zz}\end{bmatrix}\mathbf{R_I^T},\\
I_{xx}=I_z+I_y,\quad\, I_{yy}=I_z+I_x,\quad\, I_{zz}=I_x+I_y,
\end{aligned}
\end{equation}
where $\mathbf{I_{PA}}$ is the inertia tensor along the principal axes and $\mathbf{R_I}\in~$SO(3) is the rotation matrix defining their orientation, which 
can be expressed in any favorable representation e.g. Euler, Tilt, Fused angles~\cite{Allgeuer2015}.
Humans, when standing or performing locomotion tasks, such as walking or running are naturally extended in the z-axis direction due to their upright posture. 
This also has a reflection in their inertia, where the $I_{xx}$ and $I_{yy}$ principal moments are dominant~\cite{erdmann1999geometry}. Their common z-axis 
component $I_z$ is largely responsible for angular momentum associated with tilting, while $I_x$, $I_y$ for the yaw momentum $I_{\psi}$ with an angle of $\psi_I$.  
With anthropomorphic humanoid robots similar rules apply, which we use to control the orientation of the principal axes of inertia. At first, 
we model the \textit{tilting inertia} $I_z$ of the system with a simple dumbbell model. 
Two point masses $m_l,m_u$ are used to represent the lower and upper body of the robot:
\begin{equation} 
m_l = m_{ll}+m_{rl}, \qquad
m_u = m_t+m_{la}+m_{ra}.
\eqnlabel{lowerupperbodymass}
\end{equation}
The masses connected with a massless rod of length $l_I$, that tilts about an axis crossing the CoM, which in turn creates angular momentum. 
With an uneven mass distribution, the masses are spaced at varying distances $l_l,l_u$ from the CoM, which sum up to $l_I$. 
While satisfying the CoM and inertia:
 \begin{equation}
  l_u m_u = l_lm_l,
 \end{equation}
 \begin{equation}
 I_z  = m_ll_l^2 + m_ul_u^2,
 \end{equation}

\begin{figure}[!t]
\centering{\includegraphics[width=0.99\linewidth]{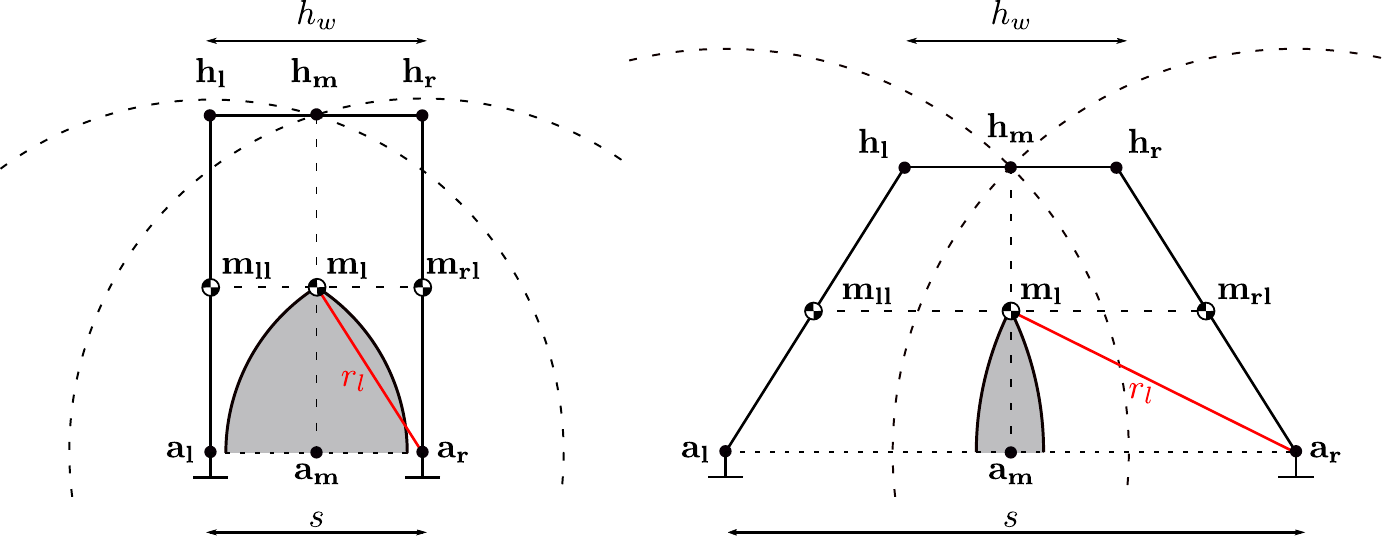}} %
\caption{Range of hip motion and resulting lower mass placement. The distance $s$ between the ankles $\mathbf{a_l,a_r}$, largely determines the possible configuration
of the robot. The grey area depicts the valid lower mass $\mathbf{m_l}$ placement region.}
\figlabel{hipext}
\vspace{-2ex}
\end{figure}
we calculate the tilting inertia as a function of $l_I$, to achieve the set distance in terms of desired inertia:
\begin{equation}%
l_I = \sqrt{I_z \frac {m_l+m_u} {m_lm_u} }.
\end{equation}
The corresponding distances are then equal to:
\begin{equation} \eqnlabel{massdistances}
l_l = l_I \frac{m_u}{m_l+m_u}, \qquad
l_u = l_I \frac{m_l}{m_l+m_u}.
\end{equation}
Finally, we compute the desired lower and upper body mass placement $\mathbf{m_l,m_u}$, which realises the set tilt inertia:
\begin{equation}\eqnlabel{massplacement}
\mathbf{m_l} = \mathbf{R_I}\begin{bmatrix}0\\0\\1\end{bmatrix}(-l_l),
\qquad \mathbf{m_u} = \mathbf{R_I}\begin{bmatrix}0\\0\\1\end{bmatrix}l_u.
\end{equation}

\subsection{Foot placement influence} \seclabel{footplacement}

Due to kinematic constraints, a whole-body pose that satisfies all of the constraints might not always be attainable. 
A large part of that is dependent on the placement of the feet. Given the desired foot placement $\Sigma_{FL},\Sigma_{FR}$, 
the left and right ankle position $\mathbf{a_l},\mathbf{a_r}$ can be computed as:
\begin{equation} 
\mathbf{a_{*}} = \mathbf{f_{*}} + \mathbf{R_{F_{*}}o_{f_{*}}},
\end{equation}
where $\mathbf{R_{F_{*}}}\in$~SO(3) is the rotation matrix with the foot orientation, 
and $\mathbf{o_{f_{*}}}$ is the position offset from the center of that foot polygon $\mathbf{f_{*}}$ to its ankle $\mathbf{a_{*}}$.
A maximum CoM extension is reached, when all of the masses are located in a single, opposite direction to the feet, with respect to kinematic constraints.
For a human or an anthropomorphic humanoid robot, it is an upright standing pose with fully stretched legs and arms extended upwards.
Naturally, the hip midpoint $\mathbf{h_m}$ is also furthest away from the feet. In terms of the dumbbell model from \secref{dumbbell}, at this full hip extension 
the maximum distance between the ankles $\mathbf{a_l},\mathbf{a_r}$ and lower body mass $\mathbf{m_l}$ is achieved. The maximum hip midpoint $\mathbf{h_m}$ 
and $\mathbf{m_l}$ range is deduced as in \figref{hipext}. From the ankles, fully extended legs connect at a distance defined by the hip width $h_w$. 
A virtual ankle midpoint $\mathbf{a_m}$ is placed on the line connecting both ankles, which are separated by distance $s$. Additionally, $\mathbf{a_h}$ 
representing the aggregate heading of the feet x-axis, is placed at an offset from $\mathbf{a_m}$.
The lower body mass then lies inside the intersection of two spheres placed at the ankles, each with a radius of $r_l$ defining the masses distance from the legs:
\begin{equation}\eqnlabel{lowermasslimit}
 \norm{\mathbf{a_*-m_l}}\leq r_l,
\end{equation}
where the value of $r_l$ is calculated simply as:
\begin{equation} \eqnlabel{radiussymbolic}
r_l = \sqrt{\left(\frac{s}{2}\right)^2+\norm{\mathbf{a_m-m_l}}_{max}^2}.
\end{equation}
The distance between the ankle midpoint and lower body mass in this case is found through a ratio of mass placement 
in terms of hip extension ($\norm{\mathbf{h_m}$ $-$ $\mathbf{a_m}}$) and a chosen leg:
\begin{equation} 
\frac{\norm{\mathbf{a_*-m_{*l}}}_{max}}{\norm{\mathbf{a_*-h_*}}_{max}} = \frac{\norm{\mathbf{a_m-m_l}}_{max}}{\norm{\mathbf{a_m-h_m}}_{max}},
\eqnlabel{legratio}
\end{equation}
where the individual lengths for a chosen leg are:
\begin{equation}
\begin{aligned} 
\norm{\mathbf{a_{*}-h_{*}}}_{max}&=c+a,\\
\norm{\mathbf{a_{*}-m_{*l}}}_{max}&=c+a-p_l(c+p_sa),\\
\norm{\mathbf{a_m-h_{m}}}_{max}&=\sqrt{(c+a)^2-\left(\frac{s-h_w}{2}\right)^2}.
\end{aligned}
\eqnlabel{leglengths}
\end{equation}     %
It can be observed, that a maximum hip extension is achieved when the separation between ankles is equal to the hip width.
Both increasing and decreasing $s$, results in a decreased range of motion for $\mathbf{h_m}$ and---as a result---$\mathbf{m_l}$.
Given a set CoM position, the orientation and amplitude of the tilting inertia is limited by the sphere intersection \eqnref{lowermasslimit}, 
as shown in \figref{inertiarange}{a}.

\begin{figure}[!t]
\centering{\includegraphics[width=0.99\linewidth]{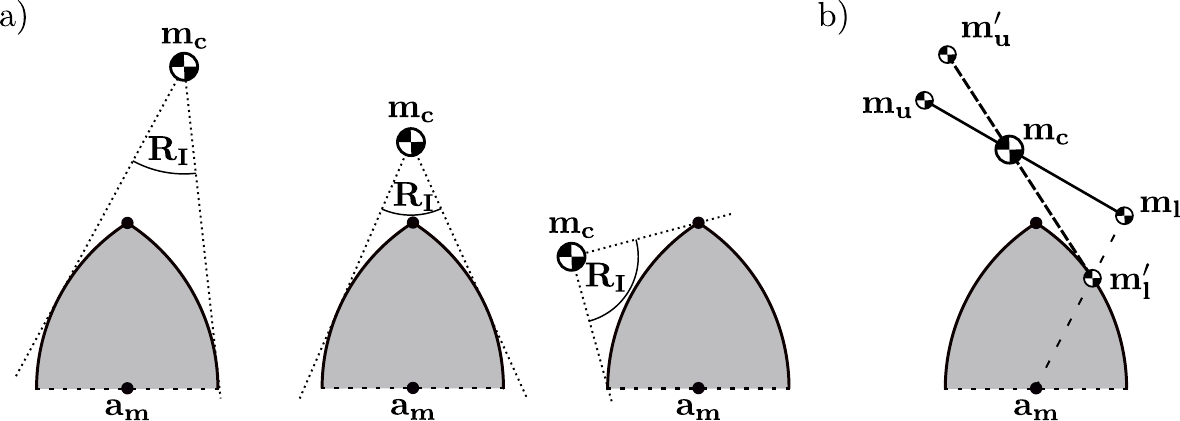}} %
\caption{The feasible inertia properties depend on the combination of foot placement and desired CoM. a) Influence of CoM on the orientation $\mathbf{R_I}$, b) preconditioning of 
unfeasible set inertia.}
\figlabel{inertiarange}
\vspace{-3ex}
\end{figure}

In case the set inertia results in $\mathbf{m_l}$ being outside of this area, we find a solution close to the desired one and recompute 
the inertial parameters. Ideally, it is possible to preserve the value of $I_z$ and tilting axis, only altering the tilt angle to achieve 
the closest point on the surface of the sphere intersection. This however, leads to discontinuous solutions for larger angles, where the closest
point might be opposite to the intended one. Instead, an intersection of a ray $\amsray{\:\mathbf{a_mm_l}\mkern-2.33mu}$
and the allowed lower mass region is used, which results in the lower mass sliding on the region's outer surface (see \figref{inertiarange}{b}). The altered rotation $\mathbf{R'_I}$ 
is then computed with \cite{MatOctRotLibGithub} as per \cite{Allgeuer2015} from the z-axis defined by the new $\mathbf{m'_l}$. Following, 
\eqnref{massdistances} is used to get $l'_I$, then $\mathbf{m'_u}$ is calculated as in \eqnref{massplacement}.

\subsection{Pose reachability} \seclabel{reachability}

The condition of the lower mass is necessary to satisfy the CoM constraint, however not always sufficient to find a valid pose, 
as the placement of $\mathbf{m'_u}$ might not be achievable. There exists a limited range how much the upper body mass can extend from the hip midpoint:
\begin{equation}\eqnlabel{distancetoupper}
d_u = \norm{\mathbf{h_m}-\mathbf{m_u}}\in[d_{min},d_{max}],
\end{equation}
where the limits are defined as a distance of the weighted trunk and arm mass, with the arms pointing towards($d_{min}$) and away($d_{min}$) 
from the hips respectively. For calculating the hip midpoint which satisfies the lower mass placement given the foot positions, 
a virtual weighted leg is created. Its parameters are linearly interpolated between the left and right leg, and the lower~($a^v$) and 
upper~($c^v$) links are shortened to accommodate for the maximum hip extension based on foot separation~\eqnref{leglengths}:
\begin{equation}
 a^v = \frac{a\norm{\mathbf{a_m-h_{m}}}_{max}}{(c+a)},\quad c^v = \frac{c\norm{\mathbf{a_m-h_{m}}}_{max}}{(c+a)}.
\end{equation}
We can then find the hip midpoint to solve \eqnref{distancetoupper}:
\begin{equation} \eqnlabel{hipmidpoint}
\mathbf{h_m} = \mathbf{p} + \mathbf{R_{R}}\left(\frac{\mathbf{a_m}-\mathbf{m_l}}{\norm{\mathbf{a_m-m_l}}}\right)(1-p^v_{si})c^v,
\end{equation}
where $\mathbf{p}$ is the point on the virtual thigh, computed as:
\begin{equation} \eqnlabel{ppoint}
\mathbf{p} = \left(\frac{\mathbf{m_l}-\mathbf{a_m}}{p^v_{li}}\right)+\mathbf{a_m},
\end{equation}
and $\mathbf{R_{R}}$ is a rotation obtained using Rodrigues' axis-angle formula. The rotation is performed at $\mathbf{p}$ around the axis $\mathbf{n}$, 
which is the normal of the virtual leg plane, defined by three points: $\mathbf{a_m}$, $\mathbf{a_{h}}$ and $\mathbf{m_l}$.
As all of the sides of the triangle formed by $\mathbf{a_m}$, $\mathbf{k}$, and $\mathbf{p}$ are of known length, the angle of 
rotation $\phi$ is calculated as the adjacent angle to $\theta$ at $\mathbf{p}$:
\begin{equation} \eqnlabel{axisangle}
\phi = \pi- \theta
\end{equation}
$$\theta = \acos\left(\frac{p^v_{li}\left(-(a^v)^2+\left(\frac{\norm{\mathbf{a_m-m_l}}}{p^v_{li}}\right) +(c^v\,p^v_{si})^2\right)}{2(c^v\,p^v_{si})\norm{\mathbf{a_m-m_l}}}\right).$$
A visual representation of this solution is shown in \figref{posereachability}. 

\begin{figure}[!t]
\centering{\includegraphics[width=0.99\linewidth]{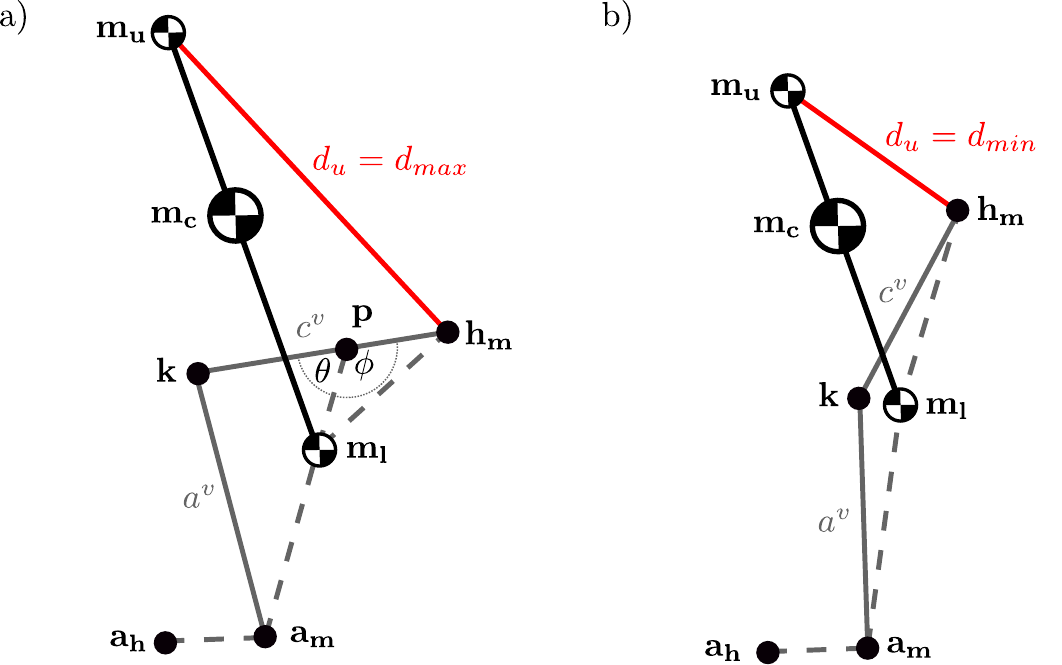}} %
\caption{Pose reachability based on upper and lower mass placement. Using a virtual single leg, the position of the hip midpoint $\mathbf{h_m}$ is calculated.
If its distance to $\mathbf{m_u}$ is evaluated to be within the permissible range of $d_u$, the pose is considered valid.}
\figlabel{posereachability}
\vspace{-3ex}
\end{figure}

Having both $\mathbf{h_m}$ and $\mathbf{m_u}$, it is necessary for the distance between them to be within range of $d_u$.
In such a case, the pose that satisfies the CoM and the tilting inertia can be reached and is considered valid and the following steps can be omitted. 
Otherwise, the tilting inertia requires further adjustment in either the orientation $\mathbf{R_{I}}$ or amplitude $I_z$.
Priority is given to keep the set orientation, as having control over it through time still allows for generating angular momentum around the CoM.
By setting $d_u$ to a value in the possible range $d_s$, leaves \eqnref{distancetoupper} then as a function of only $l_I$. Unfortunately this is an octic equation,
for which there is no general solution. However, the influence of $l_I$ on the placement of $\mathbf{m_l}$, $\mathbf{m_u}$ and $\mathbf{h_m}$ can be observed
from \eqnref{massplacement} and \eqnref{hipmidpoint}. The hip midpoint $\mathbf{h_m}$ amplifies the distance between $\mathbf{m_l}$ and $\mathbf{m_u}$ directly 
set by $l_I$. Given that the CoM placement is limited by the fully extended pose~(including the influence of foot separation $s$), there exists a $l_I$ value that
satisfies the desired $d_u$:
\begin{equation} \eqnlabel{optimisingfunction}
f(l_I) = d_u - d_s = \norm{\mathbf{h_m}-\mathbf{m_u}} - d_s,
\end{equation}
which can be found iteratively. Although the derivative of the function can be calculated for gradient-based optimisation, it is computationally 
unfavorable as it results in almost 300 operations on over 40 subexpressions, made up from 2 to 15 operations each. 
Instead, we perform a search between two $l_I$ values which produce opposite sign results on \eqnref{optimisingfunction}, as shown in \figref{optimisation}. 

\begin{figure}[!t]
\centering{\includegraphics[width=0.99\linewidth]{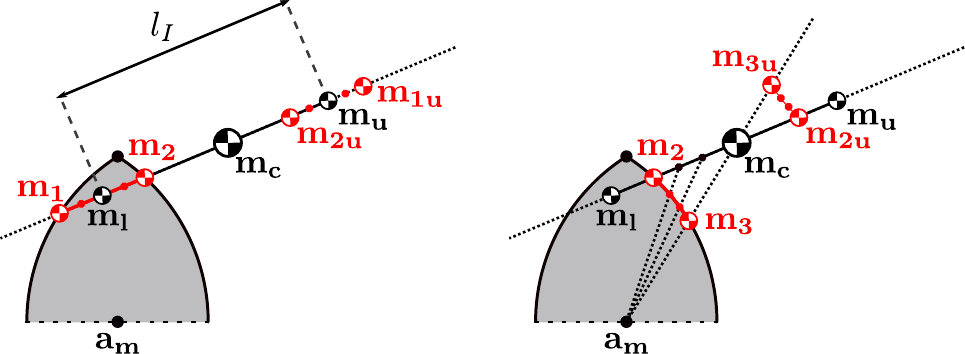}} %
\caption{A valid combination of $\mathbf{m_l}$ and $\mathbf{m_u}$ for a given CoM position lies between two search intervals of $l_I$ marked in red. 
A solution between $[\mathbf{m_1,m_2}]$, maintains CoM and $\mathbf{R_I}$. The one in $[\mathbf{m_2,m_3}]$ --- only CoM.}
\figlabel{optimisation}
\vspace{-3ex}
\end{figure}
The evaluation of the initial mass placement~\eqnref{distancetoupper} is used to obtain $d_s$. 
Either the pose is valid and $\mathbf{m_u}$ can be reached, or the hip is too close~($d_s = d_{min}$) or too far~($d_s = d_{max}$) from the upper mass.
As seen on \figref{optimisation}, three points are evaluated to assess the range of $l_I$ for the search algorithm. Points $\mathbf{m_1}$, $\mathbf{m_2}$
lie on the intersection of the tilting inertia and the lower mass placement region. If the set CoM placement is at the limit, where a fully extended pose is required, 
both $\mathbf{m_1}$ and $\mathbf{m_2}$ will be 'too far' from the solution, which requires to alter not only $l_I$, but $\mathbf{R_I}$ as well.
To assure that the CoM constraint is fulfilled, $\mathbf{m_3}$ is calculated as the intersection of the ray $\amsray{\:\mathbf{a_mm_c}\mkern-2.33mu}~(l_I=0)$ 
and the allowed lower mass region. We then compute \eqnref{optimisingfunction} for all three points, two of which are used as the $l_I$ interval.
The search is then performed with one of the various root-finding algorithms, until a sufficient accuracy has been reached e.g. bisection, regula falsi and their variants.
Searching through the range of $\mathbf{m_2}$ and $\mathbf{m_3}$~(as compared to $\mathbf{m_1}$ and $\mathbf{m_2}$) is slightly slower as it requires an additional 
recomputation of the intersection point and corresponding inertia orientation and mass placement at every step.

\subsection{Complete mass placement and CRB inertia}

With a reachable pose found that produces a given tilt inertia, the base frame position $\mathbf{h_m}$ is set. 
The base frame z-axis direction $^t\mathbf{z}$ is then computed to bring the torso mass $\mathbf{m_t}$ as close to the set upper mass $\mathbf{m_u}$,
and its yaw angle $\psi_t$ is aligned to the desired inertia yaw $\psi_I$. This completes the base frame $\Sigma_B$ definition.
Additionally, if according to \eqnref{distancetoupper} $d_u$ is far from the limits, an additional trunk tilt orientation constraint $^t\mathbf{z}_s$ can be employed.
Then, the final orientation is interpolated between $^t\mathbf{z}$ and $^t\mathbf{z}_s$, where the maximum interpolation is limited by the arm reach: $d_{max} - d_s$.

Using the base frame, we locate the hip $\mathbf{h_l}$, $\mathbf{h_r}$ and shoulder $\mathbf{s_l}$, $\mathbf{s_r}$ origins. The final leg configurations are then 
computed with the triangle approximation~\cite{ficht2018online} based on the set foot frames $\Sigma_{FL},\Sigma_{FR}$ and hip origins, which produce the leg 
mass placement $\mathbf{m_{ll}}$, $\mathbf{m_{rl}}$. 
As these masses are off the tilt axis~(separated by distance $s_l$), they contribute to the $I_x$ and $I_y$ components of the principal moments.
This is done through a combination of the lower body yaw inertia $I_l$ and the lower body yaw angle $\psi_l$, formed between the line connecting the two point masses
and the tilt inertia y-axis. A similar two-mass distribution~($\mathbf{m_{lu}}$, $\mathbf{m_{ru}}$) for the upper body ($s_u$, $\psi_u$, $I_u$) can then be used to achieve the given inertia yaw orientation, 
to generate or counteract yaw momentum. The distance and angle of the upper body mass particles around $\mathbf{m_{u}}$ can be calculated through the difference in inertia:
\begin{equation} \eqnlabel{inertiaupperdistance}
\begin{aligned}
s_u &= \sqrt{(I_\psi - I_l)\frac{m_u}{(m_{la}+\frac{m_t}{2})(m_{ra}+\frac{m_t}{2})}}&,\\
\psi_u &= \frac{(m_u+m_l)\psi_I - m_l\psi_l}{m_u}&,
\end{aligned}
\end{equation}
where $I_\psi$ is the total yaw inertia around the inertia z-axis.
Ideally, the ratio between the angles should be determined by the yaw inertia. We use the mass distribution instead---to avoid having a zero or negative value in the denominator---
when the legs provide the complete required angular momentum~($I_\psi - I_l\leq0$, $s_u=0$). To avoid self-collisions, the arm placement $\mathbf{m_{la}}$, $\mathbf{m_{ra}}$ is computed as
to keep a minimum distance to the trunk with the yaw angle $\psi_u$. Otherwise, the mass distribution is used:
\begin{equation} \eqnlabel{armplacement}
\begin{aligned}
\mathbf{m_{la}} &= \frac{\mathbf{m_{lu}}(\frac{m_t}{2}+m_{la})-\mathbf{m_{t}}(\frac{m_t}{2})}{m_{la}} &, \\
\mathbf{m_{ra}} &= \frac{\mathbf{m_{ru}}(\frac{m_t}{2}+m_{ra})-\mathbf{m_{t}}(\frac{m_t}{2})}{m_{ra}} &.
\end{aligned}
\end{equation}
A visual representation of the yaw inertia components and arm mass placement is shown on \figref{upperbodyplacement}

\begin{figure}[!t]
\centering{\includegraphics[width=0.99\linewidth]{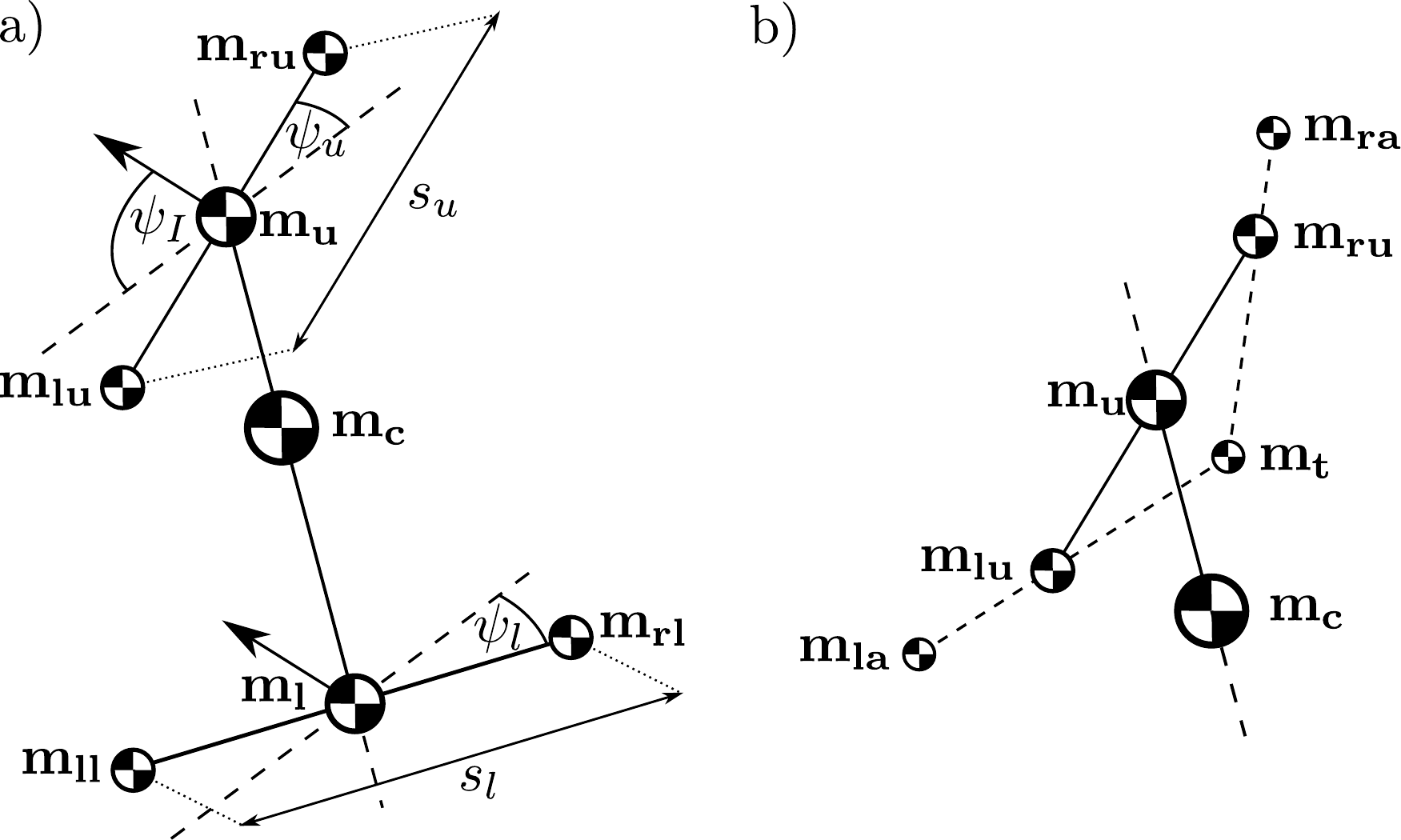}} %
\caption{Inertia yaw components and mass placement. a) achieving the desired yaw inertia with $\mathbf{m_{lu}},\mathbf{m_{ru}}$, b) completing the arm mass placement.}
\figlabel{upperbodyplacement}
\vspace{-3ex}
\end{figure}

\section{Experimental Results} \seclabel{results}

Verification of the proposed approach was performed on a \SI{90}{cm} tall, \iguhop robot~\cite{allgeuer2015child} possessing 20 position-controlled joints.
The motions were generated using the robots on-board computer.

\subsection{Modelling error}

\begin{figure}[!t]
\includegraphics[width=0.99\linewidth]{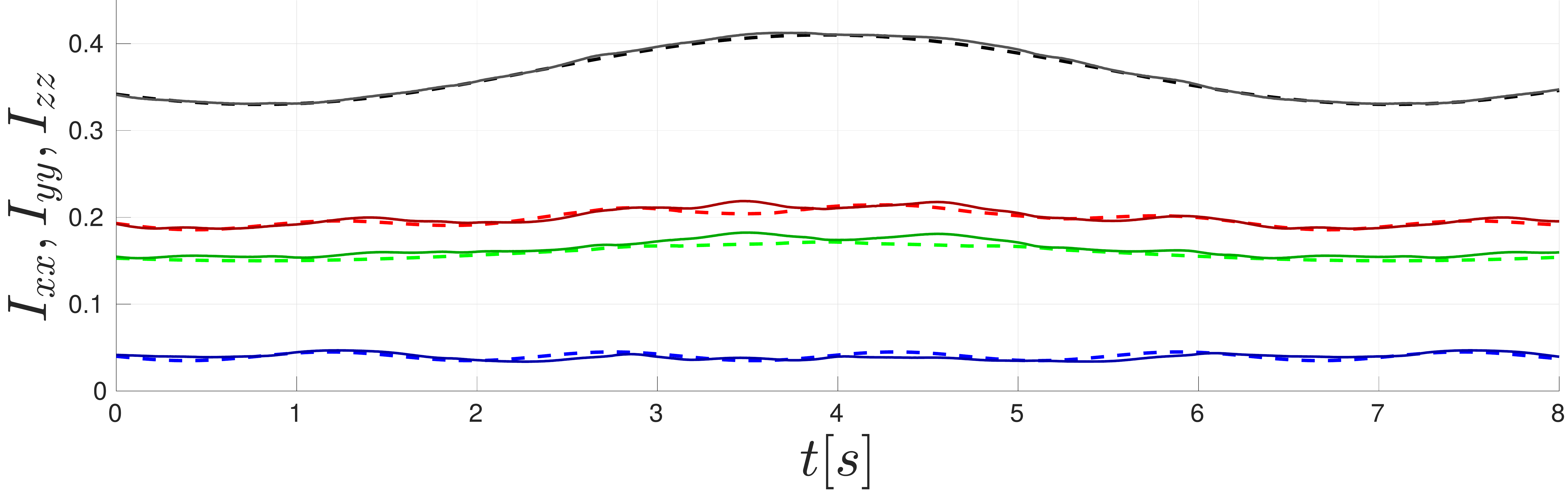}\hspace{0.01px}
\includegraphics[width=0.99\linewidth]{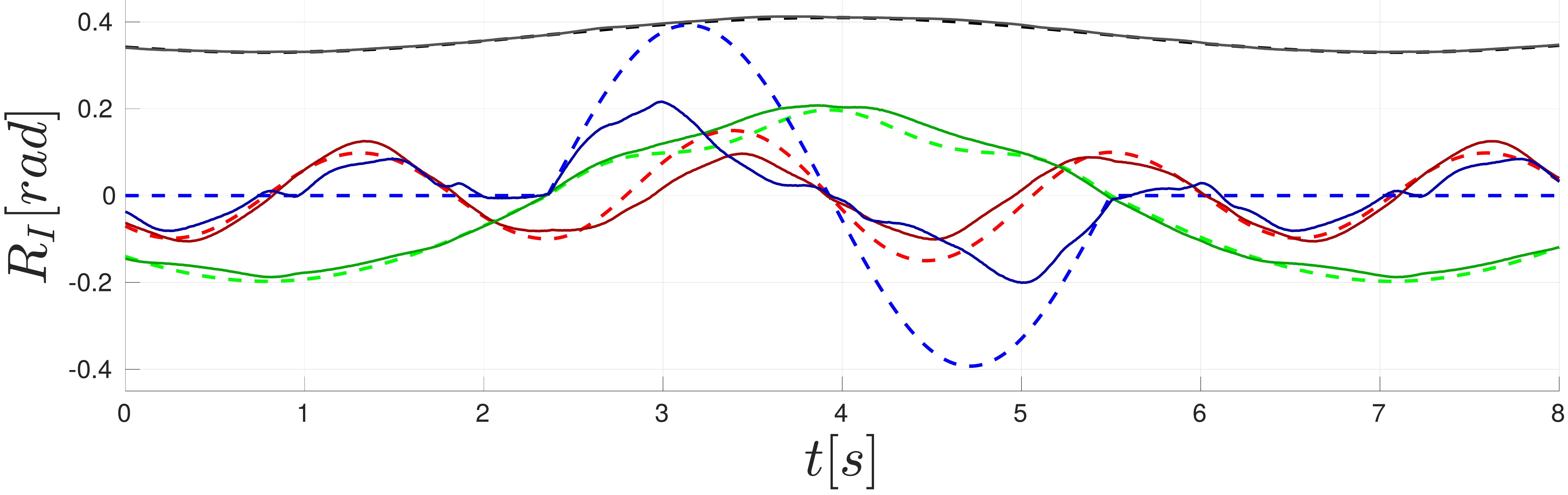}\hspace{0.01px}
\includegraphics[trim={50px 50px 584px 100px},clip,height=0.17\linewidth]{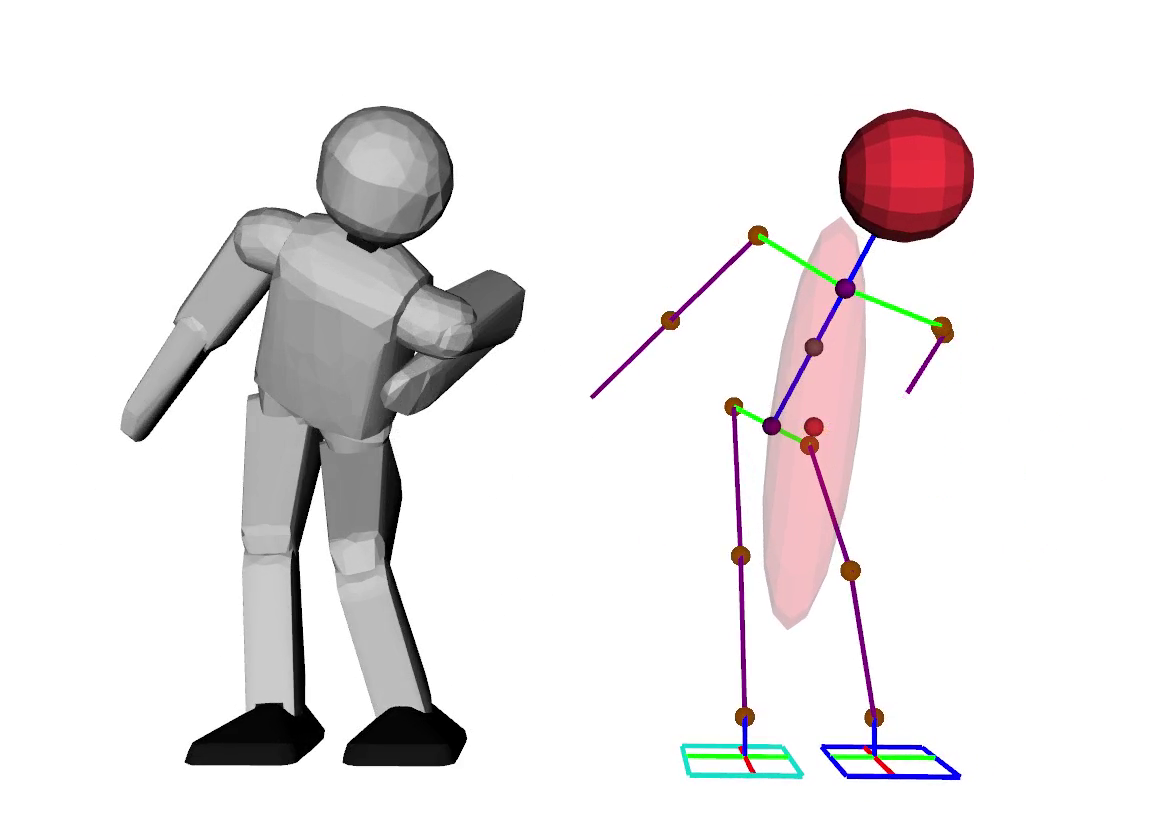}
\includegraphics[trim={50px 50px 584px 100px},clip,height=0.17\linewidth]{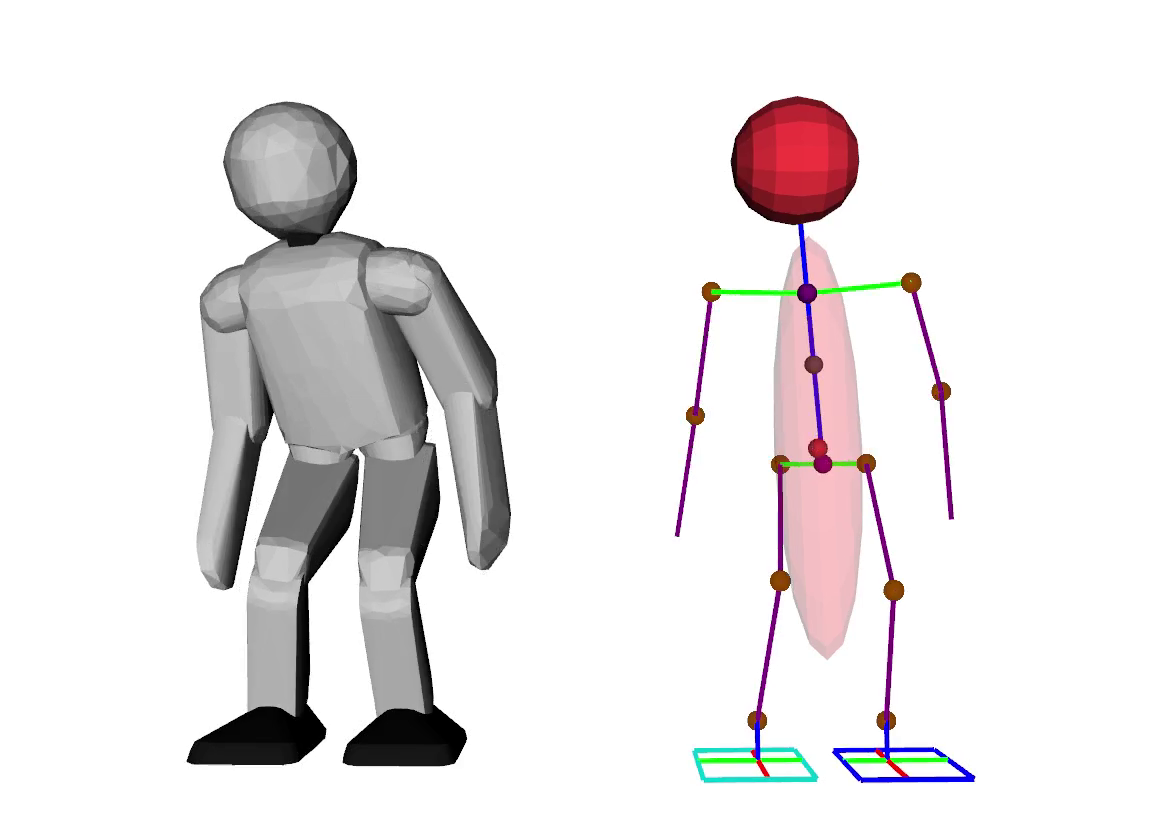}
\includegraphics[trim={90px 50px 584px 100px},clip,height=0.17\linewidth]{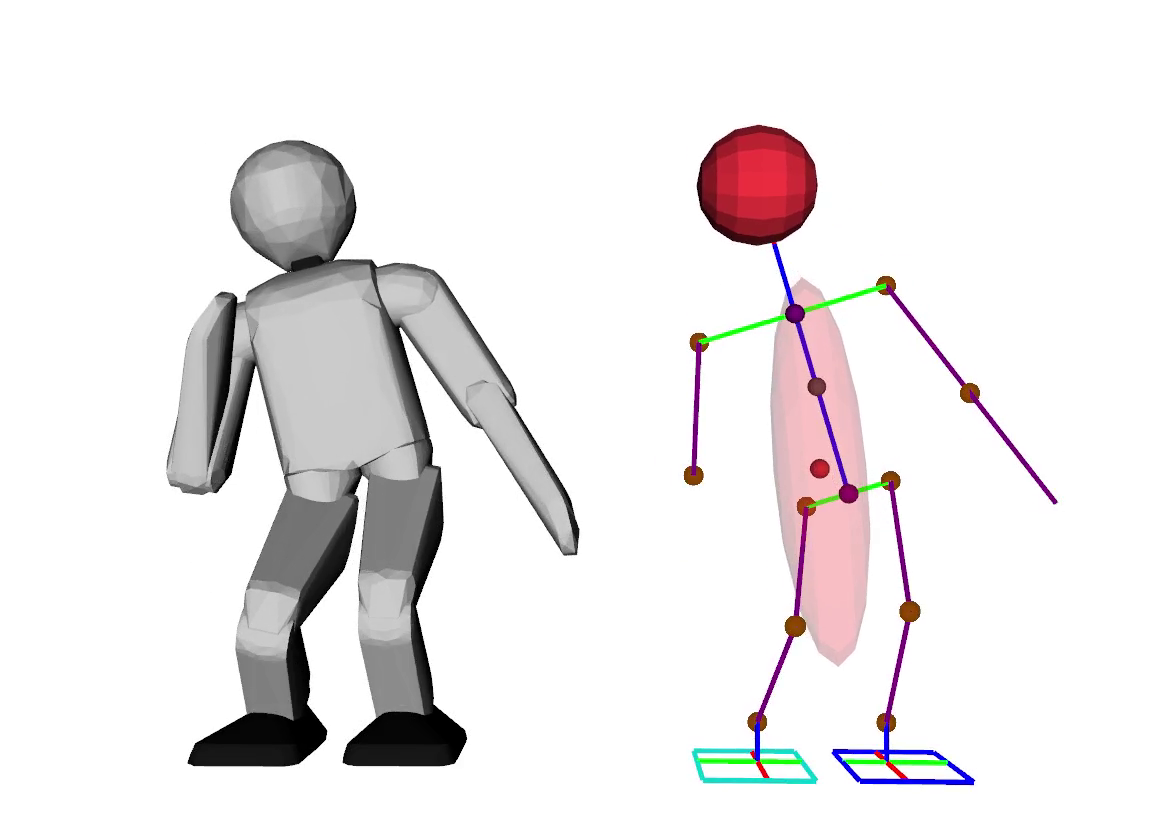}
\includegraphics[trim={50px 50px 570px 100px},clip,height=0.17\linewidth]{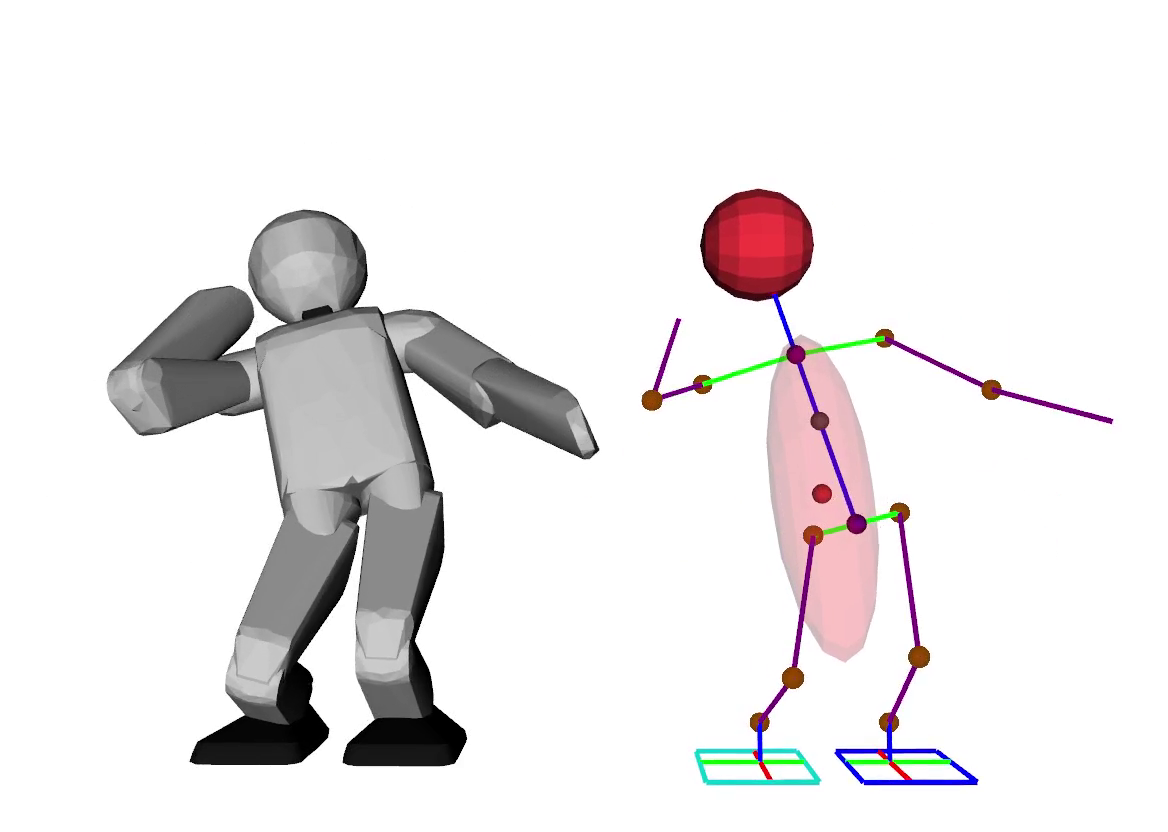}
\includegraphics[trim={50px 50px 584px 100px},clip,height=0.17\linewidth]{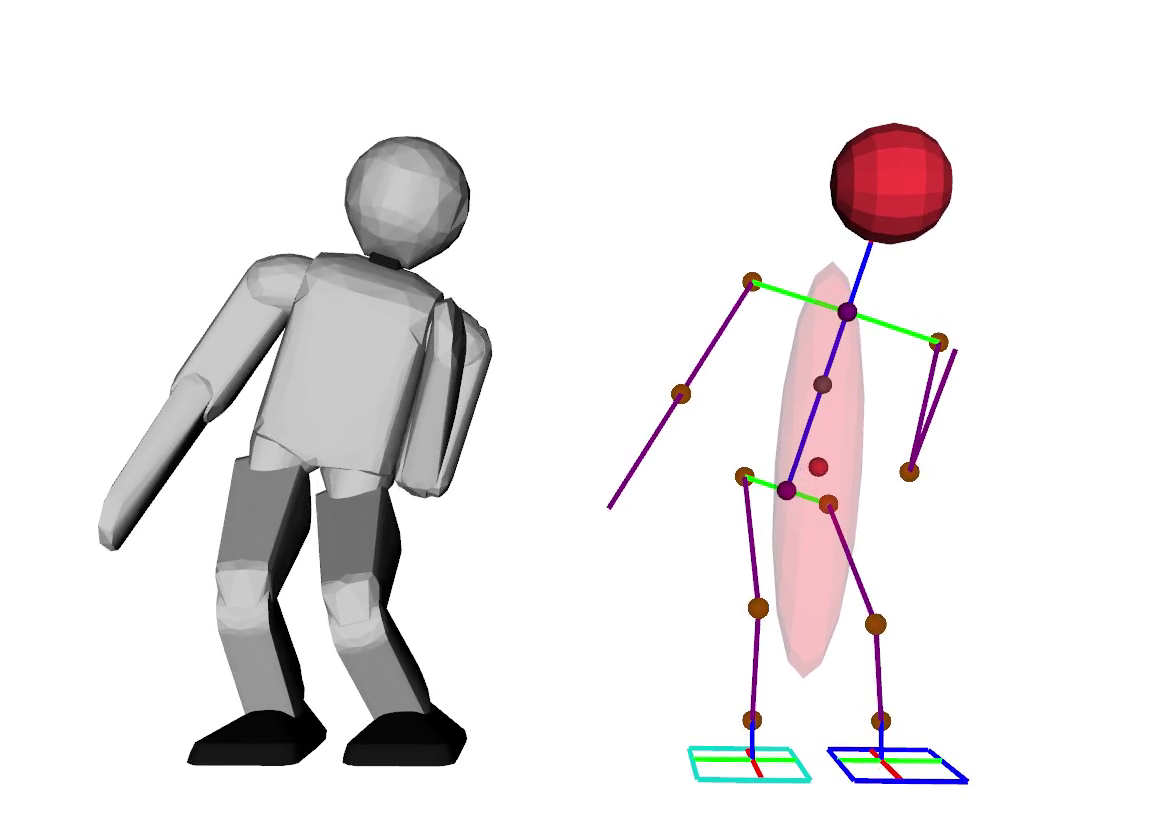}
\includegraphics[trim={80px 50px 584px 100px},clip,height=0.17\linewidth]{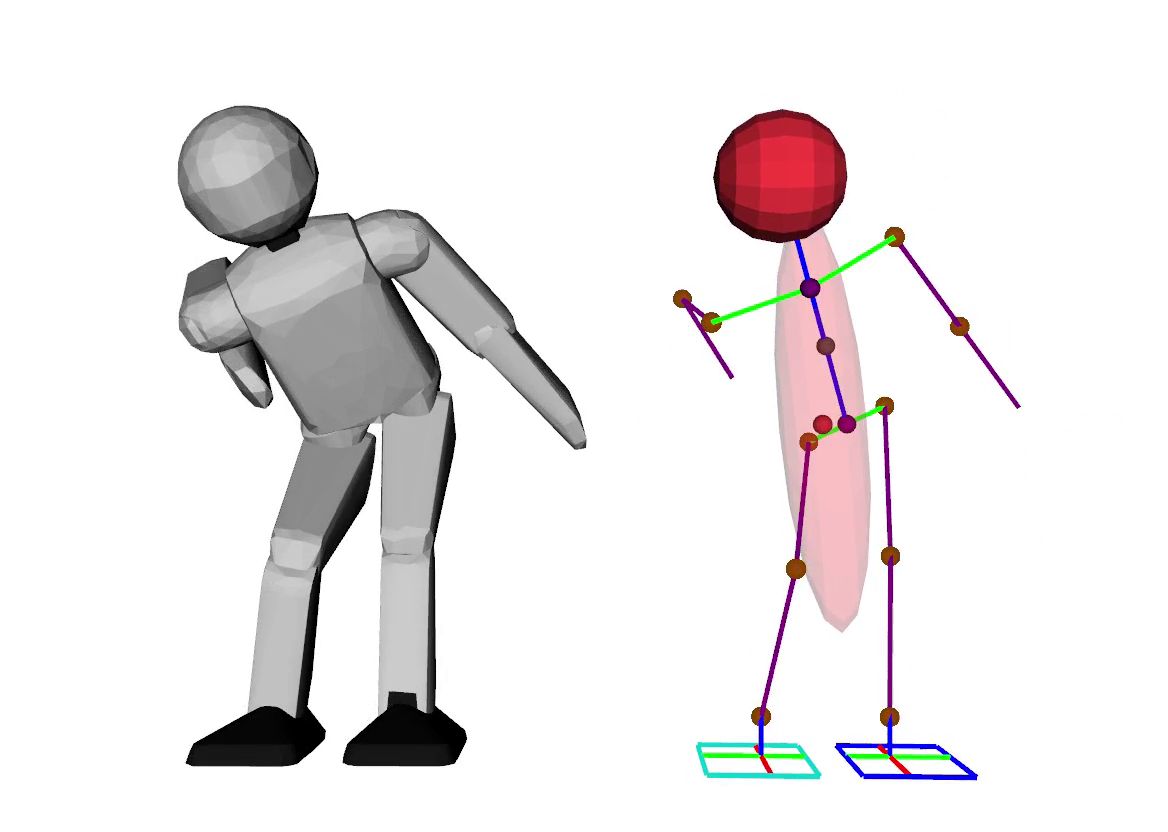}
\includegraphics[trim={50px 50px 584px 100px},clip,height=0.17\linewidth]{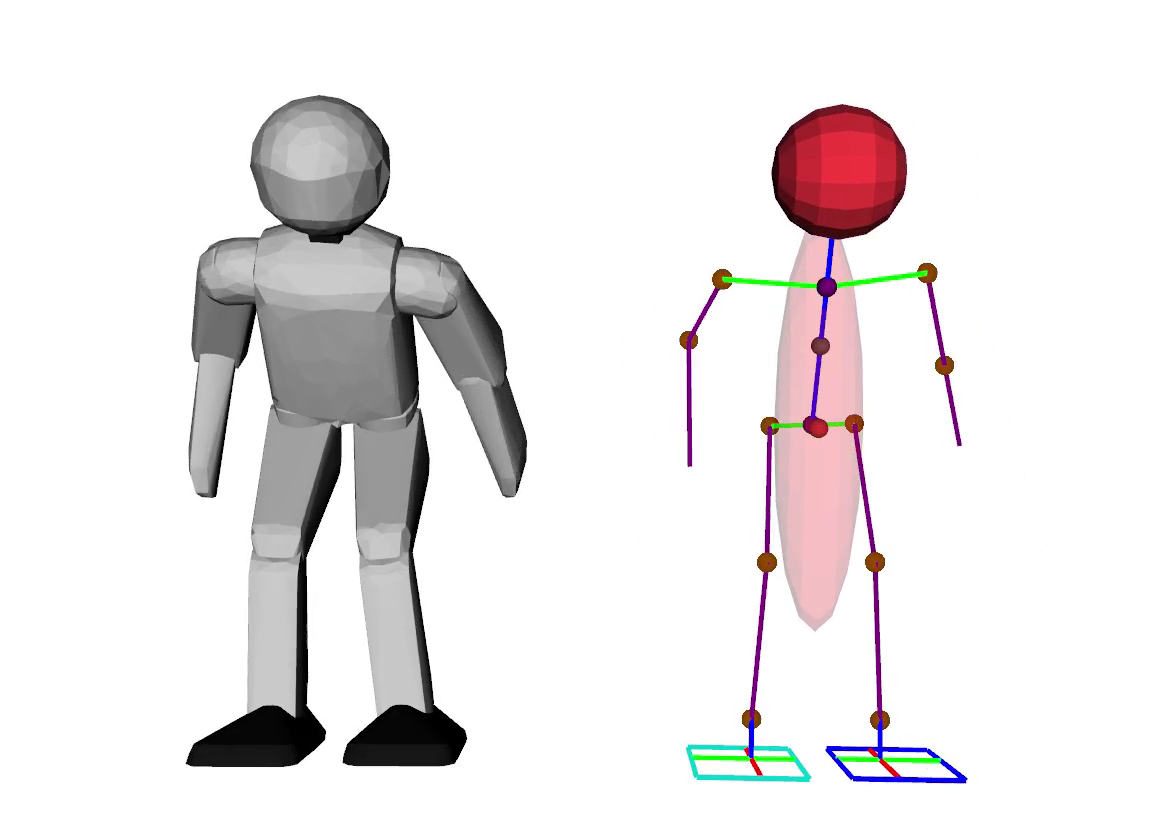}

\includegraphics[trim={584px 50px 50px 80px},clip,height=0.17\linewidth]{images/posing/0}
\includegraphics[trim={584px 50px 50px 80px},clip,height=0.17\linewidth]{images/posing/1}
\includegraphics[trim={584px 50px 50px 80px},clip,height=0.17\linewidth]{images/posing/3}
\includegraphics[trim={600px 50px 50px 80px},clip,height=0.17\linewidth]{images/posing/5}
\includegraphics[trim={584px 50px 50px 80px},clip,height=0.17\linewidth]{images/posing/7}
\includegraphics[trim={604px 50px 50px 80px},clip,height=0.17\linewidth]{images/posing/9}
\includegraphics[trim={584px 50px 50px 80px},clip,height=0.17\linewidth]{images/posing/10}
\vspace{-1ex}
\caption{Modelling error of the approach compared to a complete, precise model. From the top: Principal moments of inertia, inertia orientation, and evolution of poses 
performed by the robot over time. Dashed lines shows set values, while read values are solid. Inertia x-axis~($I_{xx}$, roll) values are red, y-axis~($I_{yy}$, pitch) green and z-axis~($I_{zz}$, yaw) blue. 
CoM height is shown in black.
The inertia is visualised as an equimomental ellipsoid~(transparent red) placed at the CoM, reflecting the mass distribution along an axis.}
\figlabel{modelerror}
\vspace{-4ex}
\end{figure}

First, we verify the validity and precision of the method by generating whole-body poses based on varying set inertial properties using the assumed model. 
The results are compared to a precise model with all masses and inertias, acting as ground truth. As seen on \figref{modelerror}, the CoM positioning 
is precise, with a mean error and standard deviation of \SI{1.5}{mm} in spite of large variations in the pose. For typical upright-scenarios, the CoM error is 
almost non-existent. The model also accurately captures the inertial properties, with small deviations in the principal moments~($I_{xx}, I_{yy}, I_{zz}$). 
The largest errors occur in the orientation $\mathbf{R_I}$. As the model consists of only point masses, the inertia products of the trunk around the 
'spine'~($\mathbf{h_m-m_t}$) are not included. Therefore, poses with a sagitally non-symmetric trunk orientation have an influence on the inertia, 
although the CoM placement is still met. This is visible in situations with~(\SI{0}{}$-$\SI{2.4}{s}) and without~(\SI{2.4}{}$-$\SI{5.5}{s}) a set yaw orientation,
where changes in the roll orientation influence the yaw and vice versa. Currently, the trunk yaw angle $\psi_t$ is aligned with the set inertia yaw 
$\psi_I$, so as to force the arms to compensate the leg movement. A revision of this strategy might provide more accurate results with respect to the 
inertia orientation. Although influenced by the trunk movement, the tilting inertia orientation is generally achieved.

\subsection{Kicking motion}

\begin{figure}[!b]
\vspace{-2ex}
\parbox{\linewidth}{\centering
\includegraphics[height=0.28\linewidth]{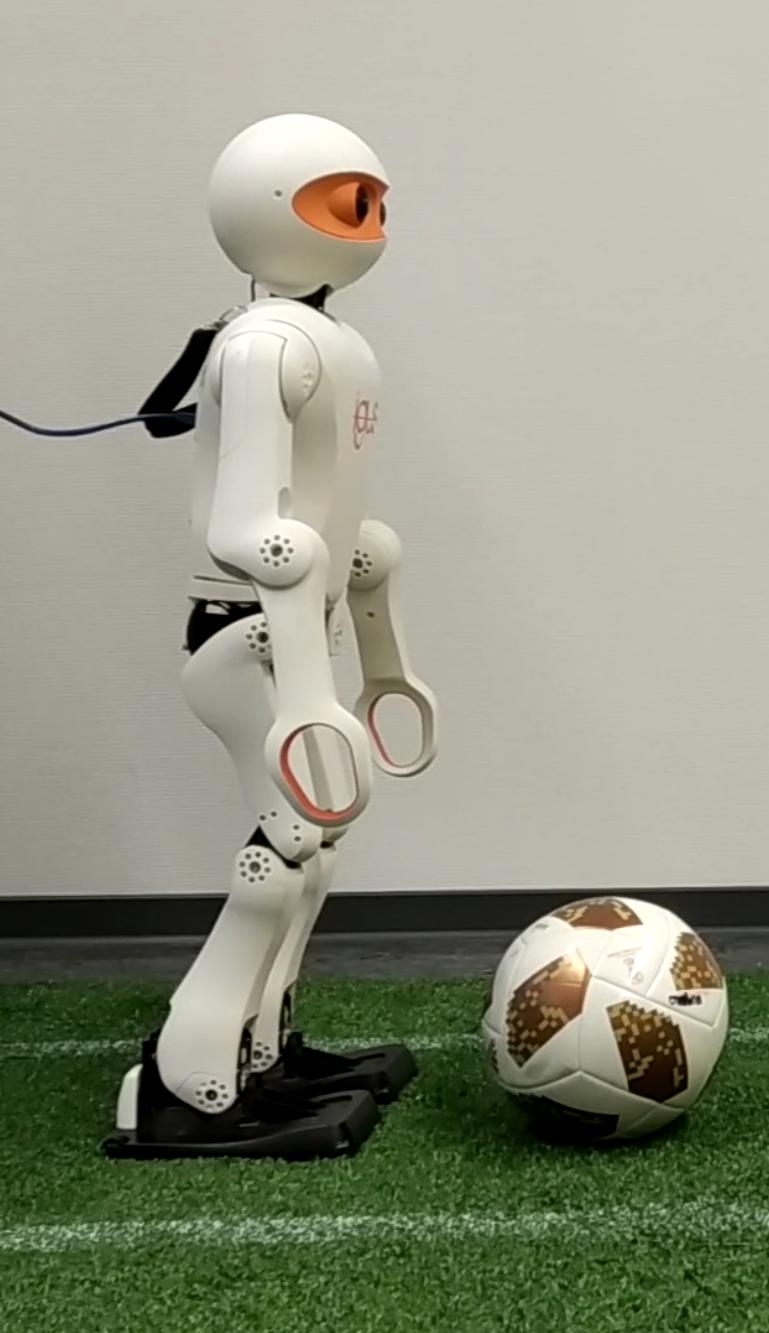}\hspace{0.01px}
\includegraphics[height=0.28\linewidth]{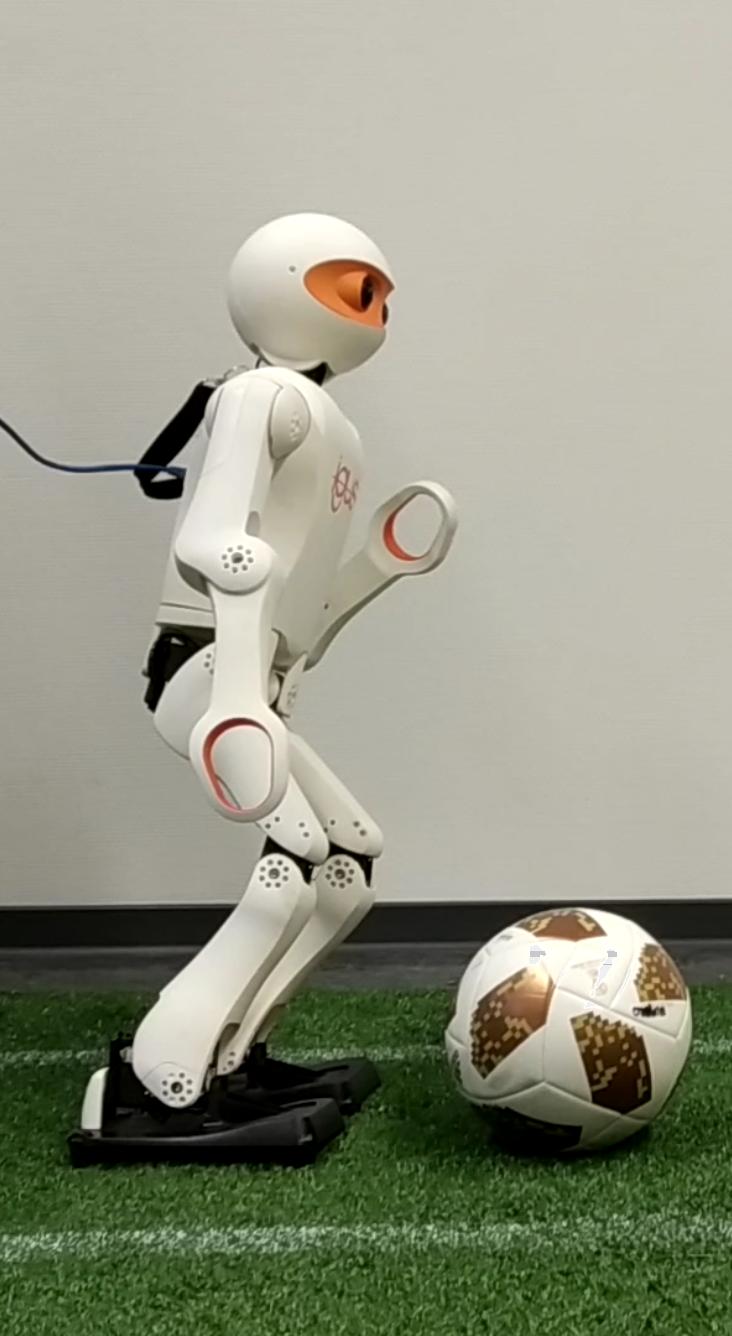}\hspace{0.01px}
\includegraphics[height=0.28\linewidth]{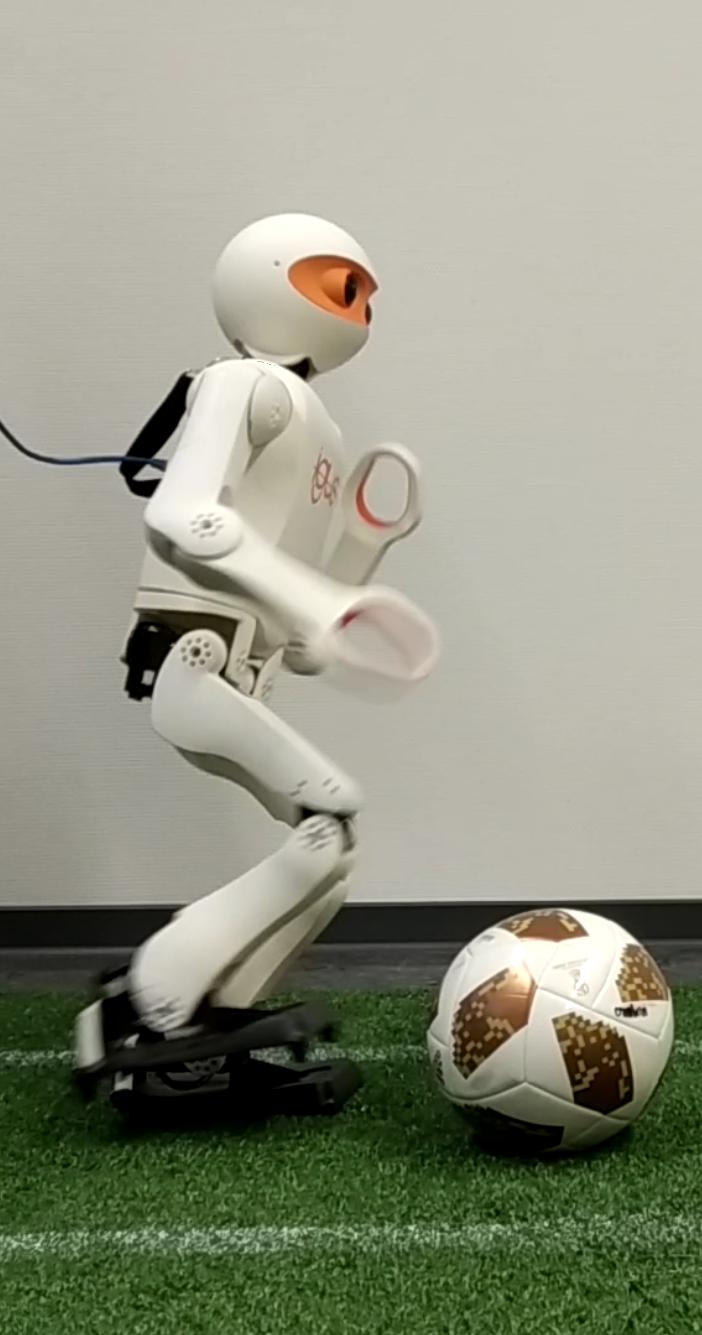}\hspace{0.01px}
\includegraphics[height=0.28\linewidth]{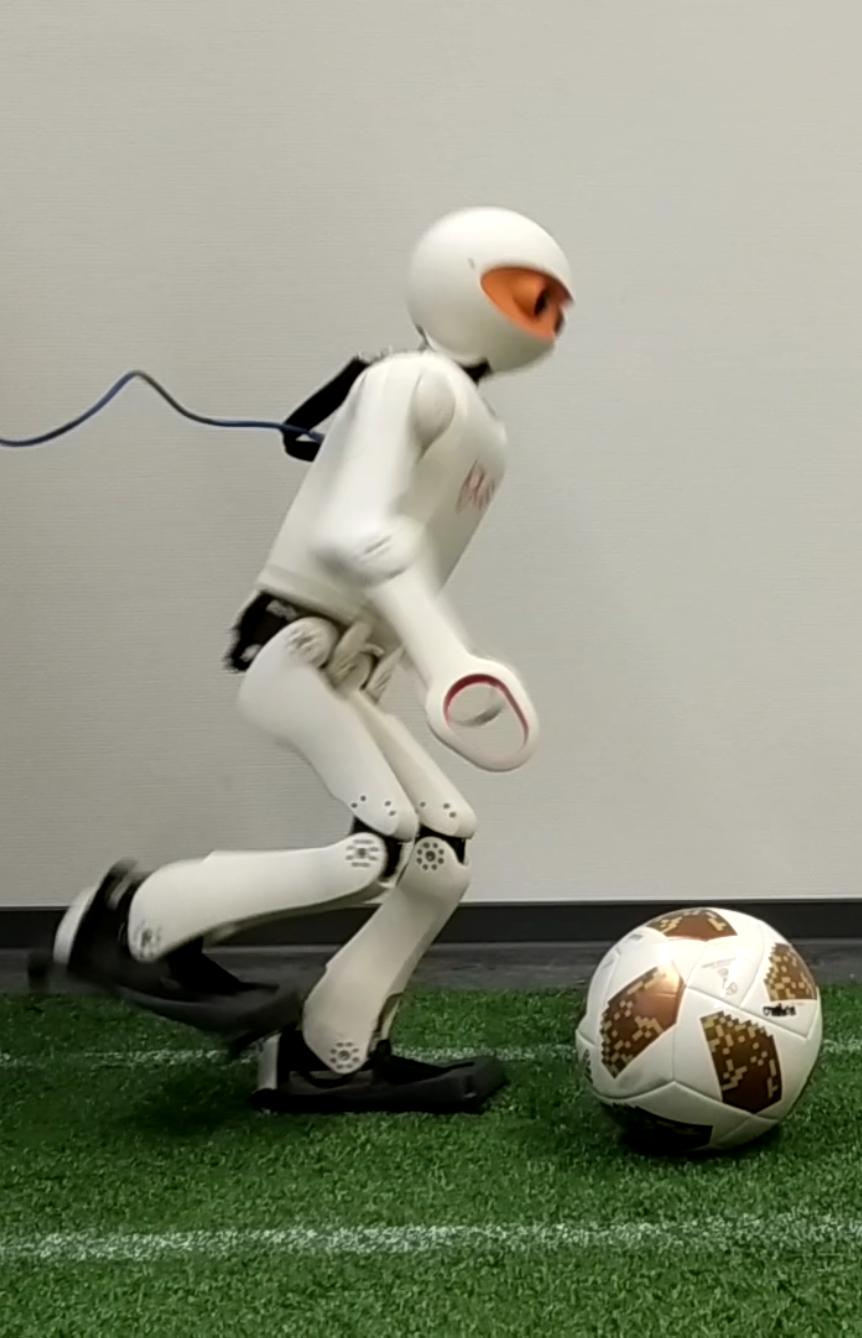}\hspace{0.01px}
\includegraphics[height=0.28\linewidth]{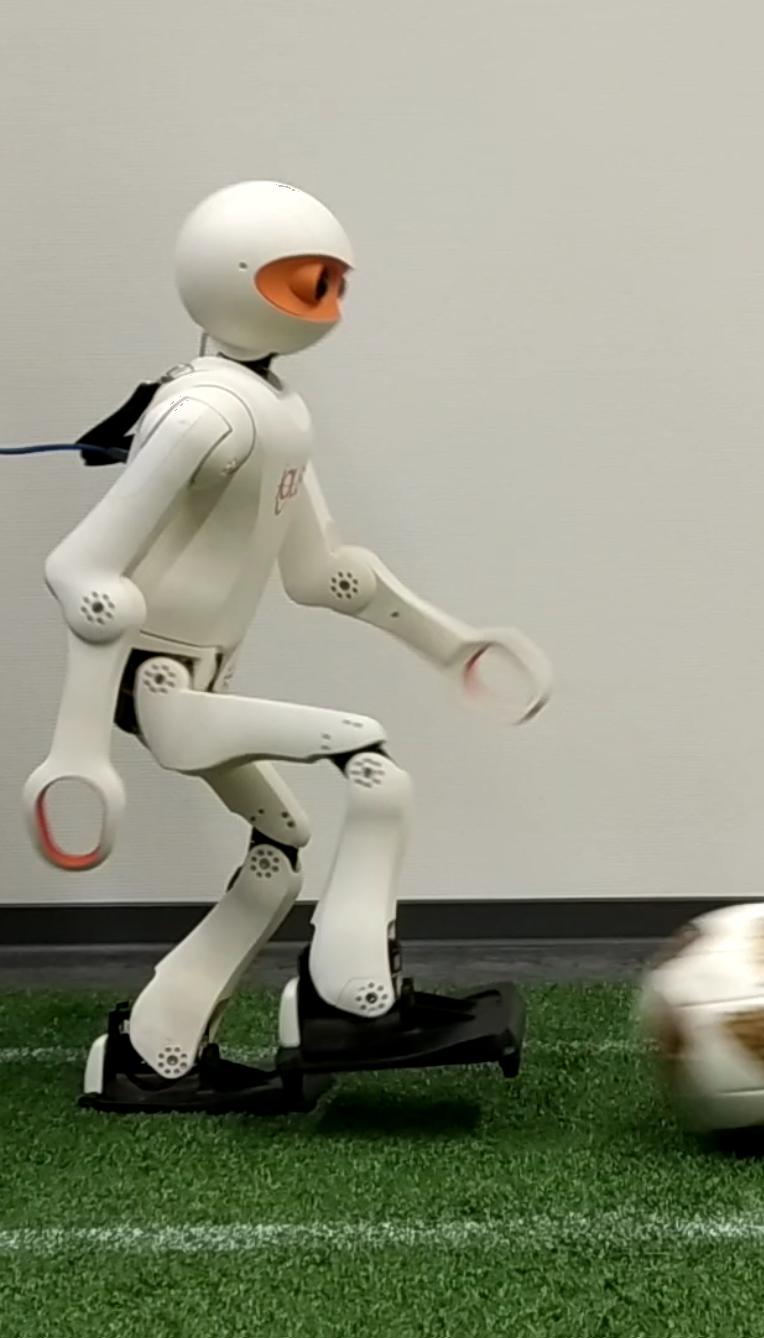}\hspace{0.01px}
\includegraphics[height=0.28\linewidth]{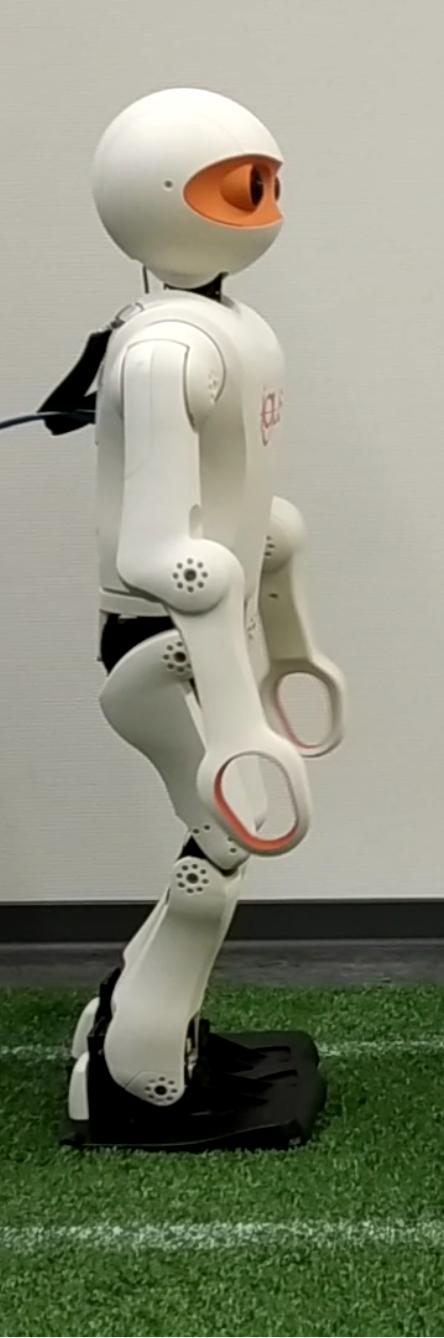}}
\caption{Still frames from a dynamic kicking motion.}
\figlabel{kick}
\vspace{-2ex}
\end{figure}

We test the motion generation by performing a dynamic kick as seen on \figref{kick}. The joint states trajectories were computed and performed on the robot on-line, using pre-designed momentum and foot frame setpoints.
The CoM is shifted above the stance foot, which enables the other foot to perform the kick, after which the robot returns to double support. During the kick, the centroidal angular momentum 
is set to zero, which results in producing vivid and natural upper body movement that counterbalances the momentum generated by the legs. Visually, the result is similar to that of the one 
generated by the Resolved Momentum Controller~\cite{kajita2003resolved}.

\subsection{Computation time}
\begin{table}[!b]
\renewcommand{\arraystretch}{1.2}
\caption{Computation time evaluation}
\tablabel{comptime}
\centering
\footnotesize
\begin{tabular}{c c}
\hline

\hline
$\qquad$Constraints met $\qquad$&$\qquad$ mean / SD ($\mu$s)$\qquad$\\
\hline
\hline
$\qquad\text{CoM}\,+\,\mathbf{R_I}\,+\,I_z$ $\qquad$&$\qquad$ 26.6475 / 3.6815$\qquad$\\
$\qquad\text{CoM}\,+\,\mathbf{R_I}$ $\qquad$&$\qquad$ 32.5346 / 7.3104$\qquad$\\
$\qquad\text{CoM}$ $\qquad$&$\qquad$ 40.6502 / 7.6436$\qquad$\\
\hline
\end{tabular}
\vspace{-1ex}
\end{table}

A common point in whole-body control is the time necessary to generate a frame of motion. We prepared three scenarios, in which the reachability of the 
pose influences the computation. The \cpp implementation of the method was executed on a single core of an Intel \mbox{i7-4710MQ} processor set to its 
base frequency of \SI{2.5}{GHz} and timed over 10000 executions. The achieved results are shown in \tabref{comptime}. On average, the algorithm requires only
33~$\mu$s to complete, which is attributed to the procedural approach using simple geometric relations. The method is the fastest when the set values can be met as it does not require to 
recompute a valid pose. For the search, regula falsi was used, which was able to achieve an error on~\eqnref{optimisingfunction} below \SI{0.1}{mm}
within 2 to 3 iterations. The quick convergence hints that~\eqnref{optimisingfunction} is mostly linear in the search interval. At the used update rate of \SI{100}{Hz},
approximately 300 motion frames can be evaluated before a single one is executed, which opens up possibilities for on-line model predictive control. 
Moreover, it is expected that a microcontroller implementation of the method running with a \SI{100}{Hz} loop is feasible, which is of importance in low-cost robotics. 

\section{Conclusions}

A novel approach to whole-body control of humanoid robots has been presented that uses a simplified non-uniform mass distribution model to produce desired 
inertial properties. The method demonstrates that control over several physical quantities of highly-dimensional and complex humanoid robots can 
be achieved with relatively simple, analytic means. As a result, fast computation times are achieved, which leaves room for integration of higher-layer 
control methods. As the proposed solution generates motions considering inertia constraints, it becomes a useful component in locomotion control schemes, 
where both linear and angular momentum are of importance. In this regard, it significantly outperforms our prior, purely CoM-based approach~\cite{ficht2018online}. 
The whole-body control has been experimentally verified and produced a naturally looking kick with a real, physical robot.

The method can be transferred to other anthropomorphic humanoids, given that the mass distribution is properly identified. This applies to biped robots as well, 
although with limitations in terms of the desired inertial properties. Other, more general task-optimising approaches can also benefit
from our results, as warm starting the solver with a quickly generated pose, can lead to faster convergence rates. In future work, we want to combine 
the whole-body controller with feedback control methods for balance in locomotion tasks.

\balance
\newpage
\bibliographystyle{IEEEtran}
\bibliography{ms}

\end{document}